\documentclass[11pt]{article}
\usepackage{amsmath}
\usepackage[preprint]{acl}

\usepackage{algorithm}
\usepackage{algorithmic}
\usepackage{times}
\usepackage{latexsym}
\usepackage{subcaption}
\usepackage{multirow}  
\usepackage{booktabs}  
\usepackage[T1]{fontenc}

\usepackage[utf8]{inputenc}

\usepackage{microtype}

\usepackage{inconsolata}

\usepackage{graphicx}
\usepackage{tabularx}
\usepackage{enumitem} 
\usepackage{tikz}
\usetikzlibrary{shapes.geometric, arrows.meta, positioning, calc, shadows, fit, backgrounds}

\title{Predictable Scaling Laws of Optimal Hyperparameters \\for LLM Continued Pre-training
}

\author{
    Yongwei Zhou\textsuperscript{1}$^{\ast}$, 
    Juncheng Diao\textsuperscript{1,2}$^{\ast}$, 
    Junlin Shang\textsuperscript{3}\thanks{Equal contribution.}, 
    Peiguang Li\textsuperscript{1},
    Rongxiang Weng\textsuperscript{1} \\
    \textsuperscript{1} MeiTuan \;\;  
    \textsuperscript{2} University of Chinese Academy of Sciences \;\;  
    \textsuperscript{3} Harbin Institute of Technology \\
    ywzhouphd2018@gmail.com \;\; diaojuncheng24@mails.ucas.ac.cn
}

\begin{document}
\maketitle

\begin{abstract}
The efficacy of continued pre-training for Large Language Models (LLMs) hinges upon hyperparameter configurations, such as learning rate and batch size. However, current practices often rely on heuristics or grid searches, leading to training instability and excessive costs. In this work, we first empirically discover that optimal hyperparameters follow stable and predictable scaling laws throughout the continued pre-training process. 
Leveraging these insights, we propose a novel framework to establish quantitative relationships between compute budget and optimal hyperparameters for a given checkpoint. Our approach has two stages: (1) \textit{Empirical Law Discovery}, where we train small-scale proxy models to derive functions mapping compute budget to optimal hyperparameters via standard loss-compute scaling laws; and (2) \textit{State-Aware Hyperparameter Prediction}, where we evaluate an initial checkpoint's validation loss and use the inverse scaling law to estimate its \textit{equivalent pre-training compute}---the compute needed to achieve the same loss from scratch. Combining this with the planned compute budget, we predict optimal hyperparameters for the target run. Empirical results demonstrate that our method reduces the hyperparameter search overhead by up to 90\% while achieving comparable or superior performance relative to baselines. This model-agnostic framework generalizes across architectures, providing a principled and efficient methodology for diverse continued pre-training scenarios starting from any given point.
\end{abstract}

\section{Introduction}

Continued pre-training (CPT) is pivotal for adapting LLMs to domain-specific applications~\cite{gupta2023continual, roziere2023code, azerbayev2024llemma}, offering a compute-efficient alternative to training from scratch. During this adaptation, LLMs encode complex knowledge structures, making them highly sensitive to hyperparameter configurations~\cite{ke2023continual}. Inappropriate learning rates ($LR$) and batch sizes ($B$) often lead to training instability or performance degradation~\cite{wen2024wsd, luo2024empirical}. However, current practices still rely on heuristics or exhaustive grid searches, which are computationally expensive and lack principled guidance~\cite{yang2022tensor, defazio2024schedule}.
\begin{figure}[t]
    \centering
    \includegraphics[width=0.48\textwidth]{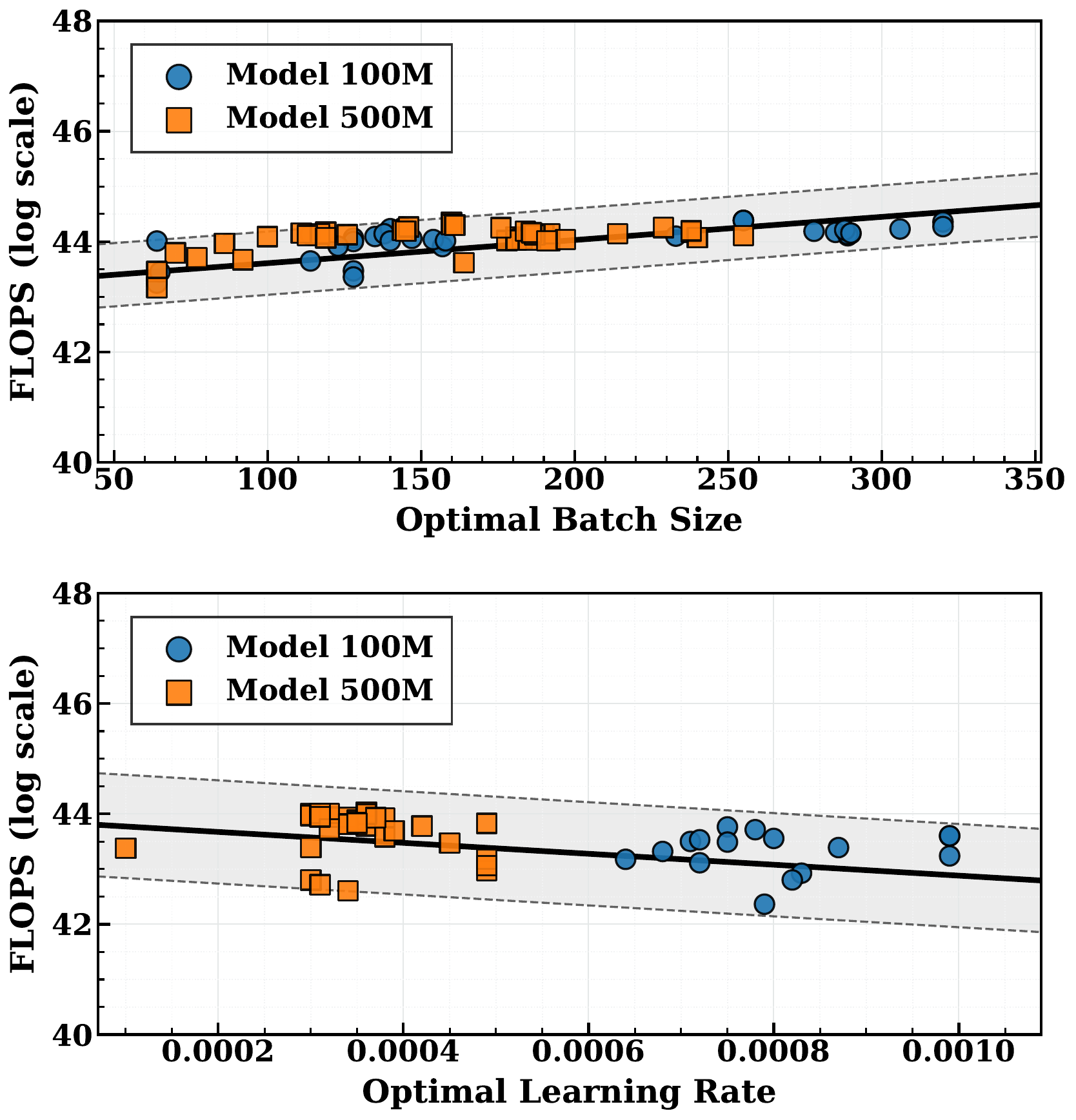}
    \caption{Scaling Laws for Optimal Hyperparameters in Continued Pre-training}
    \label{fig:hyper_scaling}
\end{figure}

While scaling laws provide such guidance for training from random initialization~\cite{kaplan2020scaling, hernandez2021scaling}, their application to CPT remains largely under-explored. Deriving scaling laws for CPT presents two core challenges: (1) it is unclear whether predictable relationships exist between model state and optimal hyperparameters when training does not start from scratch~\cite{gupta2023continual, ibrahim2024simple}; and (2) distribution shifts between pre-training and target data complicate quantifying the effective contribution of the initial checkpoint to the training process~\cite{gadre2024beyond, xie2024doremi}. In this paper, we first empirically discover that optimal hyperparameters in CPT indeed follow stable and predictable scaling laws. Leveraging this insight, we address the hyperparameter selection challenge by establishing a framework for zero-shot prediction across a given checkpoint, quantifying the relationship between compute budget and optimal configurations.

\textbf{Addressing Challenge 1: Discovering Hyperparameter Scaling Laws.} 
To investigate the existence of these laws, we systematically train small-scale proxy models across varied scales and hyperparameter configurations on CPT data. Our experiments reveal consistent hyperparameter scaling laws (Fig.~\ref{fig:hyper_scaling}): optimal batch size increases monotonically with compute budget, while optimal learning rate decreases accordingly. This empirical discovery confirms that CPT maintains stable scaling relationships—distinct from scratch-training laws—that account for the prior knowledge encoded in initial checkpoints. Based on these observations, we derive mapping functions that relate compute budget directly to optimal $LR$ and $B$ via standard loss-compute scaling laws.

\textbf{Addressing Challenge 2: Quantifying the Initial Training State.} 
To apply these laws to a arbitrary checkpoints, we introduce the concept of \textit{equivalent pre-training compute} ($C_{\text{pre}}$). Unlike training from scratch, CPT starts from a non-zero optimization state that has historically been difficult to quantify. We propose that any checkpoint can be situated as a specific coordinate on a continuous training trajectory. By mapping a model's validation loss to a compute-equivalent value on the target domain, $C_{\text{pre}}$ formally quantifies its \textit{initial training state} relative to the new task. This transforms the initial checkpoint from a ``black box'' into a measurable starting state, where the total effective compute is defined as $C_{\text{total}} = C_{\text{pre}} + C_{\text{cpt}}$, where $C_{\text{cpt}}$ is the planned CPT budget. By substituting $C_{\text{total}}$ into our derived hyperparameter scaling laws (Eqs.~\ref{eq:B_vs_C} and \ref{eq:lr_vs_C}), we achieve zero-shot prediction of the optimal learning rate and batch size for the target run.

In summary, our main contributions are threefold: (1) We establish scaling laws for optimal hyperparameters in continued pre-training, quantitatively relating compute budget to optimal learning rate and batch size, offering a principled alternative to heuristics and grid search. (2) We propose \textit{equivalent pre-training compute} to formally quantify the training state inherited from initial checkpoints. This metric transforms an arbitrary starting point into a measurable coordinate, enabling reliable hyperparameter prediction for models at any stage of their lifecycle. (3) We empirically validate our framework on models up to Dense-8B and MoE-3B parameters, demonstrating that it reduces computational costs by 70–90\% compared to traditional grid-search baselines while consistently improving training stability and performance.

\begin{figure*}[t]
    \centering
    \includegraphics[width=1\textwidth]{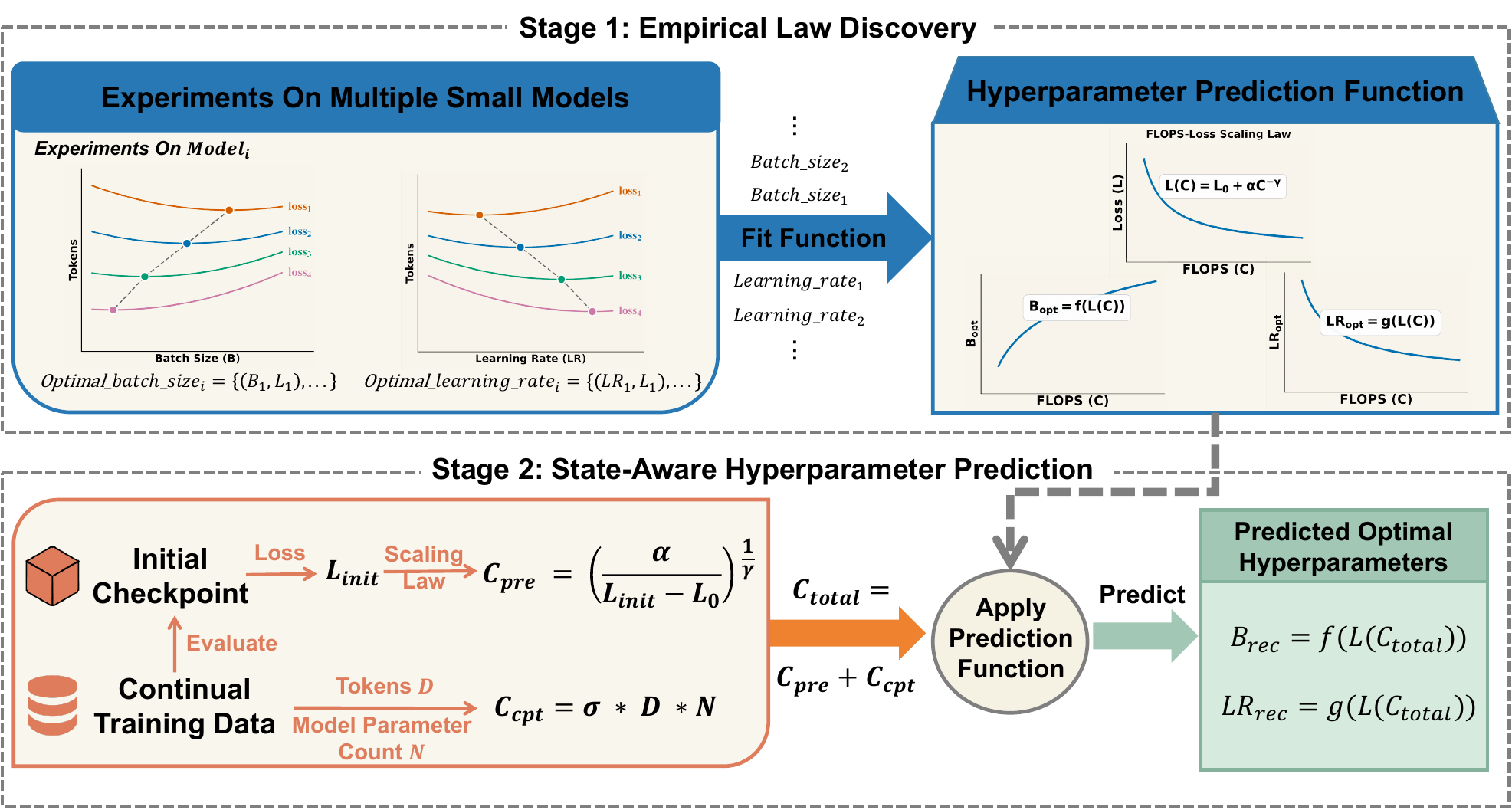}
    \caption{The Overall Framework for Optimal Hyperparameters Prediction in continued pre-training}
    \label{fig:framework}
\end{figure*}

\section{Method}

\subsection{Problem Definition and Challenges}
\label{subsec:problem}

This paper investigates the existence of \textit{hyperparameter scaling laws} that quantitatively relate the compute budget to the optimal hyperparameter configurations in continued pre-training (CPT). Establishing such laws would provide a principled framework for hyperparameter selection at any stage of a model's lifecycle, replacing computationally expensive heuristics. However, formulating these laws requires overcoming two fundamental challenges:

\begin{itemize}[leftmargin=*]
    \item \textbf{Existence and Functional Characterization:} It remains theoretically and empirically unverified whether predictable, monotonic relationships exist between the compute budget (or validation loss) and optimal hyperparameters when training does not originate from a random initialization. If such relationships exist, identifying their specific functional forms (e.g., power laws) and ensuring their stability across different model scales and architectures is a non-trivial task.

    \item \textbf{Quantifying the Initial Optimization State:} Unlike training from scratch, CPT begins from a non-zero optimization state inherited from a pre-trained checkpoint. Given that pre-training and CPT typically involve distinct data distributions, the initial checkpoint represents a ``black box'' in terms of its progress on the target domain. A core challenge lies in precisely situating this checkpoint as a measurable coordinate on a continuous training trajectory—specifically, quantifying its inherited knowledge as an equivalent compute budget relative to the target data.
\end{itemize}

\subsection{Overall Framework}
\label{subsec:framework}

To address the challenges of non-zero initialization and distribution shift, we propose a two-stage framework that models the evolution of optimal hyperparameters along a continuous optimization trajectory (Fig.~\ref{fig:framework}). This framework provides a unified approach to hyperparameter transfer, transforming the ``black-box'' initial state of any checkpoint into a quantifiable computational coordinate.

\noindent\textbf{Stage 1: Empirical Law Discovery.}
We first characterize the scaling behavior by training small-scale proxy models of varying sizes $N$ under diverse hyperparameter configurations ($LR$, $B$) on the target CPT data. By analyzing the optimal hyperparameter configurations of different validation loss, we establish two foundational components: 
(1) Hyperparameter-Loss Mappings: The functional relationships between optimal hyperparameters and the achieved validation loss, denoted as $B_{\text{opt}}=f(L)$ and $LR_{\text{opt}}=g(L)$; 
(2) Loss-Compute Scaling Laws: The standard power-law relationship $L(C)$ that describes how validation loss scales with compute budget $C$ on the target distribution.

\noindent\textbf{Stage 2: State-Aware Hyperparameter Prediction.}
Given a pre-trained checkpoint $M_{\theta_{0}}$ with an initial validation loss $L_{\text{init}}$ on the target domain, we perform a zero-shot prediction of the optimal CPT configurations. 
First, we utilize the inverse loss-compute scaling law to estimate the \textit{equivalent pre-training compute} $C_{\text{pre}} = L^{-1}(L_{\text{init}})$. This step effectively situates the checkpoint as a specific coordinate on the training trajectory. 
Second, we define the \textit{total effective compute} as $C_{\text{total}} = C_{\text{pre}} + C_{\text{cpt}}$, where $C_{\text{cpt}}$ is the planned CPT budget. 
Finally, the optimal hyperparameters for the target run are predicted by evaluating $f(C_{\text{total}})$ and $g(C_{\text{total}})$ (or equivalently $f(L_{target})$ and $g(L_{target})$).

\subsection{Empirical Law Discovery}
\label{subsec:coldstart}

This stage establishes the foundational mappings between validation loss $L$ and the optimal hyperparameter configurations: $B_{\text{opt}}(L)$ and $LR_{\text{opt}}(L)$. Since loss $L$ itself follows a predictable scaling law with compute budget $C$ (i.e., $L = \mathcal{S}(C)$), these relationships implicitly define the trajectory of optimal hyperparameters as compute scales: $B_{\text{opt}} = f(\mathcal{S}(C))$ and $LR_{\text{opt}} = g(\mathcal{S}(C))$. 

\paragraph{Definition of Optimality.}
For a given model scale and dataset, we define the optimal configuration $(B_{\text{opt}}, LR_{\text{opt}})$ for a target loss $L$ as the one that reaches the target loss with the minimum training compute. Formally, among all configurations $\mathcal{H} = \{(B, LR)\}$ that achieve a validation loss $\leq L$, $(B_{\text{opt}}, LR_{\text{opt}})$ is optimal if it minimizes the total compute $C$ (measured in FLOPs), thereby identifying the most compute-efficient path to a specific optimization state.

\paragraph{Empirical Data Collection via Proxy Models.}
To discover these laws, we utilize small-scale proxy models ($N \in \{100\text{M}, 500\text{M}\}$) trained on the target CPT dataset. For each scale, we execute an extensive grid search over $B$ and $LR$, monitoring validation loss across a multi-domain benchmark (General, Math, Code). The resulting loss-versus-compute trajectories for each $(N, B, LR)$ triplet provide the empirical basis for our scaling analysis.

\paragraph{Three-Step Derivation Process.}
We derive the mapping function through the following steps:

\textbf{Step 1: Constructing Iso-Loss Curves.}
For a target loss level $\hat{L}$, we identify the required data tokens $\hat{D}$ for each hyperparameter configuration. By plotting these estimates in the $(B, LR)$ plane, we construct iso-loss curves as illustrated in Fig.~\ref{fig:loss_lr_batch}, where each point on a curve corresponds to the number of data tokens $\hat{D}$ required to reach the target loss $\hat{L}$ on a specific hyperparameter configuration. The vertex of each curve—representing the minimum $\hat{D}$ to reach $\hat{L}$—uniquely identifies the optimal pair $(B_{\text{opt}}(\hat{L}), LR_{\text{opt}}(\hat{L}))$.

\textbf{Step 2: Functional Fitting.}
We observe that the relationship between loss and optimal hyperparameters follows a power-law trend. We fit the collected pairs $\{(\hat{L}_i, B_{\text{opt}, i})\}$ and $\{(\hat{L}_i, LR_{\text{opt}, i})\}$ to the following parametric forms:
\begin{equation}
B_{\text{opt}}(L) = f(L) \quad \text{and} \quad LR_{\text{opt}}(L) = g(L).
\end{equation}

\textbf{Step 3: Composition with Compute Scaling Laws.}
Following \citet{kaplan2020scaling}, we model the loss-compute relationship as:
\begin{equation}
L(C) = L_0 + \alpha \cdot C^{-\gamma},
\label{eq:scaling_law}
\end{equation}
where $L_0$ is the irreducible loss, and $\gamma$ is the scaling exponent. By substituting Eq.~\ref{eq:scaling_law} into our fitted functions, we obtain the final Compute-to-Hyperparameter Scaling Laws:
\begin{align}
B_{\text{opt}}(C) &= f(L_0 + \alpha \cdot C^{-\gamma}), \label{eq:B_vs_C} \\
LR_{\text{opt}}(C) &= g(L_0 + \alpha \cdot C^{-\gamma}). \label{eq:lr_vs_C}
\end{align}

These laws enable zero-shot prediction of optimal hyperparameters for any target compute budget $C_{\text{total}}$, circumventing the need for further tuning.

\subsection{State-Aware Hyperparameter Prediction}
\label{subsec:warmstart}

Given an initial pre-trained checkpoint $M_{\theta_{0}}$ and a target CPT compute budget $C_{\text{cpt}}$, this stage predicts the optimal hyperparameters by situating the checkpoint on the previously derived scaling trajectory. The core challenge lies in quantifying the optimization state of a pre-trained model when it is introduced to a new data distribution.

Our framework rests on the hypothesis that a pre-trained model's state on a target domain can be uniquely characterized by an \textit{equivalent pre-training compute} ($C_{\text{pre}}$). We define $C_{\text{pre}}$ as the theoretical compute budget required to reach the model's current validation loss if it had been trained from a random initialization on the target data. This hypothesis allows us to treat CPT not as a disjoint training phase, but as a continuation of a single, contiguous optimization path. We empirically validate this hypothesis across diverse domain-shift scenarios in Section~\ref{sec:experiments}.

\textbf{Step 1: Projecting the Initial State.}
We first evaluate the checkpoint's initial validation loss $L_{\text{init}}$ on the target CPT data. By inverting the loss-compute scaling law $L(C) = L_{0} + \alpha C^{-\gamma}$ established in Section~\ref{subsec:coldstart}, we calculate the \textit{Equivalent Compute}:
\begin{equation}
C_{\text{pre}} = \left( \frac{\alpha}{L_{\text{init}} - L_{0}} \right)^{1/\gamma}.
\label{eq:C_pre_calc}
\end{equation}
This step effectively projects the "black-box" checkpoint onto the target domain's compute-loss coordinate system.

\textbf{Step 2: Total Effective Computation.}
The total effective compute budget for the CPT phase is defined as the sum of the inherited and planned compute:
\begin{equation}
C_{\text{total}} = C_{\text{pre}} + C_{\text{cpt}}.
\label{eq:C_total}
\end{equation}
This formulation re-centers the CPT process onto the global scaling curve, where $C_{\text{total}}$ represents the terminal point of the intended training trajectory.

\textbf{Step 3: Optimal Hyperparameter Prediction.}
We first predict the terminal validation loss $L_{\text{target}}$ at the end of the CPT run: $L_{\text{target}} = L(C_{\text{total}})$. The optimal hyperparameters are then predicted by evaluating the optimality functions at the predicted state:
\begin{align}
B_{\text{opt}} &= f\left(L_0 + \alpha \cdot C_{\text{total}}^{-\gamma}\right), \label{eq:B_opt}\\
LR_{\text{opt}} &= g\left(L_0 + \alpha \cdot C_{\text{total}}^{-\gamma}\right).
\end{align}

The complete procedure is summarized in Algorithm~\ref{alg:ohp_framework} (Appendix~\ref{sec:algorithm}). This approach provides a principled, zero-shot mechanism for hyperparameter selection, directly addressing the non-zero initialization challenge via the $C_{\text{pre}}$ formulation.

\section{Experiments}
\label{sec:experiments}

We evaluate the proposed framework through a series of large-scale experiments designed to answer three key research questions: 
\textbf{RQ1 (Predictability):} Do optimal hyperparameters in CPT follow stable scaling relationships across different scales? 
\textbf{RQ2 (Accuracy):} Can the \textit{Equivalent Compute} $C_{\text{pre}}$ accurately quantify the initial state of a given checkpoint under domain shift? 
\textbf{RQ3 (Efficiency):} Does our framework consistently outperform heuristic baselines while reducing the total search cost?

\subsection{Datasets and Evaluation Protocol}
\label{subsec:datasets}

To test the robustness of our framework across diverse data distributions, we curate a heterogeneous CPT corpus spanning three distinct domains: general knowledge, mathematics, and formal code. 

\paragraph{Training Corpus.} 
As shown in Table~\ref{tab:exp_dataset}, we sample approximately 55B tokens from high-quality open-source repositories: SlimPajama~\cite{weber2024redpajama} for general linguistic knowledge, OpenWebMath~\cite{paster2023openwebmath} for mathematical reasoning, and GitHub for programming logic. 
To ensure unbiased evaluation, we construct a held-out validation set by randomly sampling 0.5\% of documents from each domain. 

\begin{table}[t]
    \centering
    \small
    \caption{Composition of the 55B Token Continued Pre-training Corpus.}
    \label{tab:exp_dataset}
    \begin{tabular}{l l c c}
        \toprule
        \textbf{Domain} & \textbf{Source Dataset} & \textbf{Tokens} & \textbf{Ratio} \\
        \midrule
        General    & SlimPajama   & 45.0B & 82.0\% \\
        Math       & OpenWebMath  & 5.0B  & 9.0\%  \\
        Code       & GitHub       & 5.0B  & 9.0\%  \\
        \midrule
        \textbf{Total} & --           & \textbf{55.0B} & \textbf{100\%} \\
        \bottomrule
    \end{tabular}
\end{table}

\paragraph{Evaluation Benchmarks.}
To comprehensively evaluate model performance and the effectiveness of the predicted hyperparameters,
we use a diverse suite of downstream benchmarks.
\textbf{(1)~Knowledge:}
MMLU~\cite{hendrycks2020measuring} and MMLU-Pro~\cite{wang2024mmlu} for general world knowledge.
\textbf{(2)~Mathematical Reasoning:}
GSM8K~\cite{cobbe2021training} and MathQA~\cite{amini2019mathqa} for multi-step arithmetic and mathematical problem-solving.
\textbf{(3)~Commonsense Reasoning:}
HellaSwag~\cite{zellers2019hellaswag}, PIQA~\cite{bisk2020piqa}, WinoGrande~\cite{sakaguchi2021winogrande}, and OpenBookQA~\cite{mihaylov2018can} for physical and logical commonsense.
\textbf{(4)~Code Generation:}
HumanEval~\cite{chen2021evaluating} and LiveCodeBench~\cite{jain2024livecodebench} to evaluate functional accuracy in code generation.

\begin{figure*}[t]
    \centering
    \includegraphics[width=0.95\textwidth]{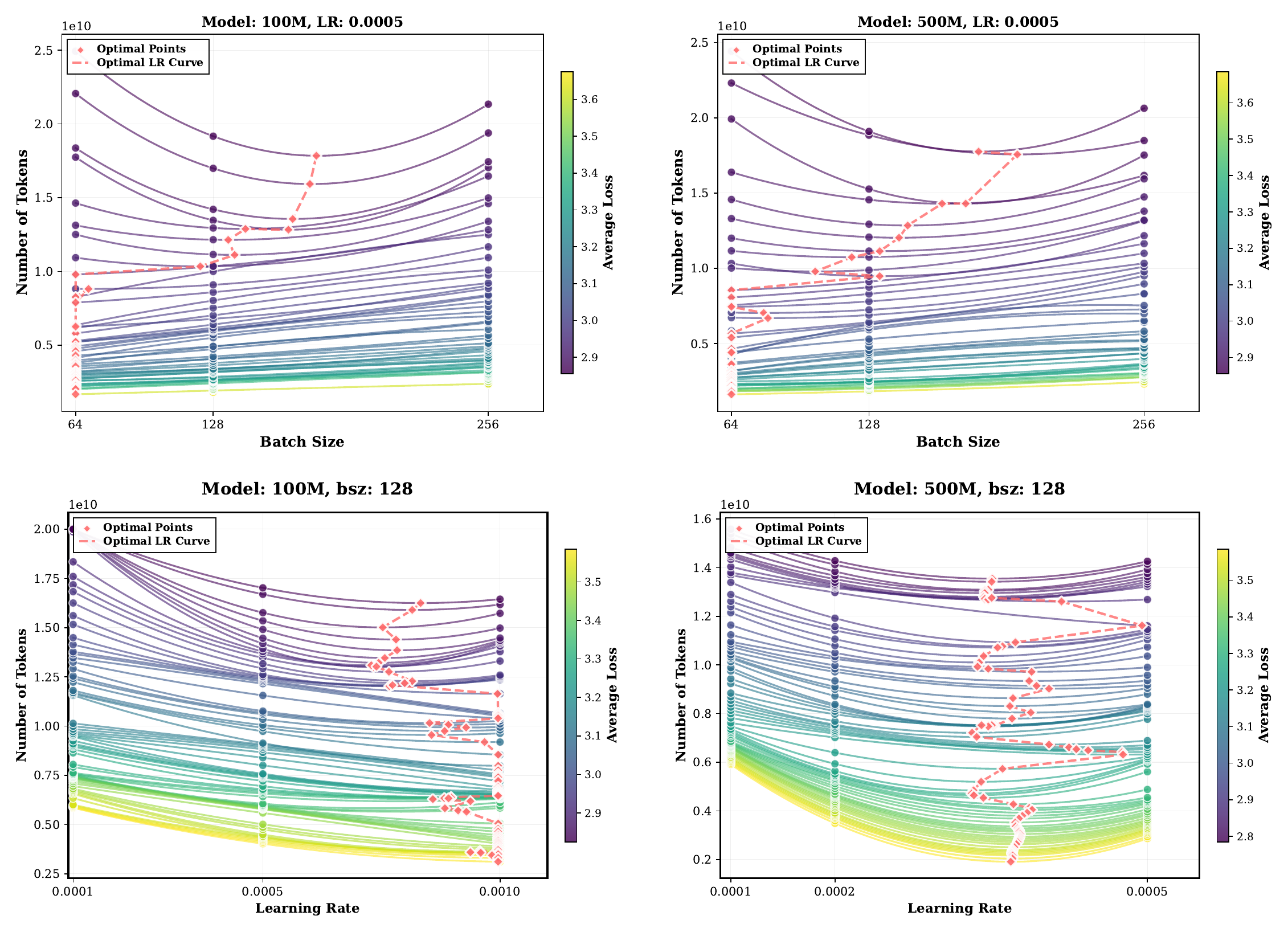}
    \caption{Optimization analysis of learning rate and batch size }
    \label{fig:loss_lr_batch}
\end{figure*}

\subsection{Settings for Empirical Law Discovery}
\label{subsec:cold_start_settings}

In this phase, we investigate the relationship between compute budget and optimal hyperparameters—specifically learning rate ($LR$) and batch size ($B$)—using the CPT data described in Section~\ref{subsec:datasets}. We evaluate both dense and Mixture-of-Experts (MoE) architectures; detailed configurations are provided in Appendix~\ref{appendix:model_details}.

For both architectures, models are trained at two distinct scales: $N \in \{100\text{M}, 500\text{M}\}$, where $N$ denotes the number of activated parameters during forward propagation. To identify optimal hyperparameter configurations at these scales, we conduct an exhaustive grid search for each of the four model variants: Dense-100M, Dense-500M, MoE-100M, and MoE-500M. The search space is defined by the following grid: $B \in \{64, 128, 256\}$ and $LR \in \{1\times 10^{-4}, 2\times 10^{-4}, 5\times 10^{-4}\}$. All experiments employ the AdamW optimizer~\cite{loshchilov2017decoupled} with its default hyperparameters.

As shown in Fig.~\ref{fig:loss_lr_batch}, the optimal $LR$ tends to decrease as the compute budget increases, whereas the optimal $B$ exhibits a corresponding increasing trend. These relationships are captured quantitatively via the scaling laws derived earlier and are illustrated in Fig.~\ref{fig:hyper_scaling}. Furthermore, Appendix~\ref{appendix:empirical_trends} details the empirical evolution of these hyperparameters, demonstrating why larger $B$ and smaller $LR$ are necessary near convergence.

\subsection{Validation Settings for Continued Pre-training}
\label{subsec:continued_pre-training_settings}

To evaluate the predictive accuracy of our framework, we employ two large-scale pre-trained checkpoints as starting points: a Dense-8B model (pre-trained on 6T tokens) and an MoE-3B model (pre-trained on 5T tokens). 

As detailed in Table~\ref{tab:continued_training_setup}, we first estimate the equivalent pre-training compute ($C_{\text{pre}}$) for each checkpoint by evaluating its initial validation loss on the target CPT corpus. This value is then aggregated with the planned compute budget for the 55B-token CPT phase ($C_{\text{cpt}}$). The resulting total effective compute budget, $C_{\text{total}} = C_{\text{pre}} + C_{\text{cpt}}$, is subsequently used to predict the optimal hyperparameters for each CPT run, leveraging the scaling laws established during the hyperparameter prediction phase (Section~\ref{subsec:coldstart}). Detailed quantitative results and the specific parameters of these scaling laws are provided in Appendix~\ref{appendix:scaling_laws}.

\begin{table}[t]
    \centering
    \caption{Experimental setup and predicted optimal ranges for continued pre-training.}
    \label{tab:continued_training_setup}
    \setlength{\tabcolsep}{3pt} 
    \resizebox{\columnwidth}{!}{%
    \begin{tabular}{@{}l c c c c c c c@{}}
        \toprule
        \multirow{2}{*}{\textbf{Model}} & \multirow{2}{*}{\textbf{Val Loss}} & \multirow{2}{*}{\textbf{$C_{\text{pre}}$}} & \multirow{2}{*}{\textbf{$C_{\text{cpt}}$}} & \multicolumn{2}{c}{\textbf{LR ($10^{-4}$)}} & \multicolumn{2}{c}{\textbf{Batch Size ($B$)}} \\
        \cmidrule(lr){5-6} \cmidrule(l){7-8}
         & & & & \textbf{Predict} & \textbf{Range} & \textbf{Predict} & \textbf{Range} \\
        \midrule
        \textbf{Dense-8B}  & 2.08 & $7.0 \times 10^{20}$ & $2.1 \times 10^{21}$ & 0.5 & [0.3, 1.0] & 768  & [703, 844] \\
        \addlinespace
        \textbf{MoE-3B}    & 1.25 & $1.2 \times 10^{21}$ & $6.7 \times 10^{20}$ & 1.0 & [0.7, 1.2] & 1600 & [1495, 1929] \\
        \bottomrule
    \end{tabular}%
    }
\end{table}

\begin{table*}[htbp]
\centering
\small
\renewcommand{\arraystretch}{1.2}
\setlength{\tabcolsep}{4pt}
\caption{Ablation results on MoE-3B and Dense-8B comparing our predicted setting against variations in $LR$ and $B$. Best results are in \textbf{bold}, and second-best are \underline{underlined}. We exclude the LiveCodeBench results for Dense-8B, as the benchmark's high difficulty leads to marginal performance and limited discriminative power.}
\label{tab:combined-results}
\resizebox{\textwidth}{!}{%
\begin{tabular}{l cc cc cccc cc c}
\toprule
\multirow{2}{*}{\textbf{Configuration}} 
& \multicolumn{2}{c}{\textbf{Math}} 
& \multicolumn{2}{c}{\textbf{Knowledge}} 
& \multicolumn{4}{c}{\textbf{Commonsense Reasoning}} 
& \multicolumn{2}{c}{\textbf{Code}} 
& \multirow{2}{*}{\textbf{Avg}} \\
\cmidrule(lr){2-3} \cmidrule(lr){4-5} \cmidrule(lr){6-9} \cmidrule(lr){10-11}
& \textbf{GSM8K} & \textbf{MathQA} & \textbf{MMLU-Pro} & \textbf{MMLU} & \textbf{HellaSwag} & \textbf{PIQA} & \textbf{Wino.} & \textbf{OBQA} & \textbf{HumanEval} & \textbf{LiveCode} & \\
\midrule

\multicolumn{12}{l}{\textbf{MoE-3B Model Settings}} \\
\midrule
\multicolumn{12}{l}{\textit{Varying Learning Rate ($B = 1600$)}} \\
$LR = 6 \times 10^{-5}, B = 1600$ & \underline{68.8} & \textbf{51.5} & 40.5 & \textbf{71.3} & 63.0 & \underline{79.5} & 63.5 & \underline{81.5} & \textbf{46.9} & \underline{9.9} & 57.6 \\
$LR = 5 \times 10^{-4}, B = 1600$ & 54.0 & 44.0 & 33.5 & 66.3 & 49.5 & \underline{79.5} & 62.0 & 74.5 & 35.4 & 9.2 & 50.8 \\
\midrule
\multicolumn{12}{l}{\textit{Varying Batch Size ($LR = 1 \times 10^{-4}$)}} \\
$LR = 1 \times 10^{-4}, B = 512$  & 67.4 & 49.5 & 39.8 & 69.0 & \textbf{70.0} & 76.0 & \underline{64.0} & 78.5 & 39.6 & \underline{9.9} & 56.4 \\
$LR = 1 \times 10^{-4}, B = 3200$ & \textbf{70.8} & \underline{51.0} & \textbf{41.2} & \underline{71.2} & \underline{66.0} & 78.0 & 63.0 & 80.5 & 45.7 & 9.5 & \underline{57.7} \\
\midrule
\textbf{Predicted ($LR = 1 \times 10^{-4}, B = 1600$)} & 66.7 & \underline{51.0} & \underline{40.6} & 70.1 & 64.0 & \textbf{81.0} & \textbf{65.0} & \textbf{82.0} & \underline{46.3} & \textbf{12.1} & \textbf{57.9} \\

\midrule \midrule 

\multicolumn{12}{l}{\textbf{Dense-8B Model Settings}} \\
\midrule
\multicolumn{12}{l}{\textit{Varying Learning Rate ($B = 768$)}} \\
$LR = 3 \times 10^{-5}, B = 768$ & \underline{40.9} & 33.2 & \textbf{26.7} & 59.8 & \textbf{77.3} & 79.2 & 71.2 & \underline{43.4} & 26.8 & - & 51.0 \\
$LR = 5 \times 10^{-4}, B = 768$ & 31.1 & 32.5 & 24.4 & 53.8 & 75.5 & 78.5 & \textbf{72.7} & 41.6 & 18.3 & - & 47.6 \\
\midrule
\multicolumn{12}{l}{\textit{Varying Batch Size ($LR = 5 \times 10^{-5}$)}} \\
$LR = 5 \times 10^{-5}, B = 256$  & 39.0 & 32.6 & 25.2 & 60.1 & 76.9 & \textbf{79.4} & \underline{71.8} & \textbf{44.6} & \textbf{29.9} & - & 51.0 \\
$LR = 5 \times 10^{-5}, B = 1600$ & \textbf{42.1} & \underline{33.6} & 26.1 & \textbf{60.4} & 77.0 & \textbf{79.4} & \underline{71.8} & 43.3 & 26.2 & - & \underline{51.1} \\
\midrule
\textbf{Predicted ($LR = 5 \times 10^{-5}, B = 768$)} & 40.7 & \textbf{33.8} & \underline{26.5} & \underline{60.3} & \underline{77.2} & \underline{79.3} & 71.0 & 42.8 & \underline{29.8} & - & \textbf{51.3} \\

\bottomrule
\end{tabular}
}
\end{table*}

\begin{table}[htbp]
\centering
\small
\renewcommand{\arraystretch}{1.2}
\setlength{\tabcolsep}{4pt}
\caption{Comparison of computational costs for hyperparameter search.}
\label{tab:compute_cost}
\resizebox{\columnwidth}{!}{%
\begin{tabular}{ll ccc}
\toprule
\textbf{Model} & \textbf{Strategy} & \textbf{Proxy Models} & \textbf{FLOPs} & \textbf{Savings} \\
\midrule
\multirow{2}{*}{\textbf{Dense-8B}} 
& Grid Search       & $9 \times \text{Dense-8B}$  & $1.02 \times 10^{22}$ & -- \\
& Ours (Scaling)    & $9 \times \{100\text{M}, 500\text{M}\}$ & $8.15 \times 10^{20}$ & \textbf{92.0\%} \\
\midrule
\multirow{2}{*}{\textbf{MoE-3B}} 
& Grid Search       & $9 \times \text{MoE-3B}$ & $3.03 \times 10^{21}$ & -- \\
& Ours (Scaling)    & $9 \times \{100\text{M}, 500\text{M}\}$ & $8.15 \times 10^{20}$ & \textbf{73.1\%} \\
\bottomrule
\end{tabular}%
}
\end{table}

\subsection{Results Analysis}
Applying our prediction protocol, the derived mapping functions yield the following optimal configurations: $\{LR=5 \times 10^{-5}, B=768\}$ for Dense-8B and $\{LR=1 \times 10^{-4}, B=1600\}$ for MoE-3B. To rigorously validate the optimality of these predictions, we systematically decouple the two hyperparameters to conduct controlled ablations. As shown in Table~\ref{tab:combined-results}, models trained with our predicted configurations consistently achieve peak average scores (51.3 and 57.9, respectively) compared to all off-target settings, empirically validating the framework's effectiveness across diverse architectures.
\paragraph{Impact of Learning Rate at Fixed Batch Size.}
With the batch size fixed at the predicted optimum, we observe that the learning rate is critical for balancing knowledge adaptation and stability. An excessively high $LR$ ($5 \times 10^{-4}$) leads to a sharp performance decline to 47.6 for Dense-8B and 50.8 for MoE-3B. This collapse, particularly pronounced in reasoning tasks such as GSM8K, suggests that excessive step sizes may disrupt pre-trained weight structures and trigger catastrophic forgetting. Conversely, smaller $LR$ (e.g., $3 \times 10^{-5}$ for Dense-8B) yield suboptimal results (51.0), as they constrain the efficiency of new knowledge acquisition within the fixed token budget. Our predicted $LR$ optimally balances preserved prior capacity against full domain knowledge learning.

\paragraph{Impact of Batch Size at Fixed Learning Rate.}
Varying the $B$ while fixing the optimal $LR$ reveals a distinct inverted U-shaped performance curve. Smaller batch sizes (256 for Dense-8B, 512 for MoE-3B) suffer from high gradient variance and unstable optimization, resulting in lower scores (51.0 and 56.4). However, increasing $B$ beyond the predicted optimum is not always beneficial. Given a fixed compute budget, an excessively large $B$ (1600 for Dense-8B, 3200 for MoE-3B) significantly reduces the total number of gradient updates, hindering the model's convergence on the target distribution. This results in suboptimal scores of 51.1 and 57.7. Our proposed $B$ balances gradient fidelity and update frequency to maximize model performance.

\subsection{Computational Efficiency Analysis}
\label{subsec:efficiency}

Compared to traditional grid searches conducted directly on large-scale target models, our framework significantly enhances computational efficiency by leveraging small-scale proxy models. As shown in Table~\ref{tab:compute_cost}, this approach reduces the total compute budget by 92.0\% for Dense-8B and 73.1\% for MoE-3B.
Specifically, a traditional grid search evaluating 9 configurations on the target models would require $1.02 \times 10^{22}$ FLOPs for Dense-8B and $3.03 \times 10^{21}$ FLOPs for MoE-3B. In contrast, our framework only performs these evaluations on 100M and 500M proxy models. Due to the substantially lower compute requirements of these proxies, our total search budget is merely $8.15 \times 10^{20}$ FLOPs. Because the search cost is largely decoupled from the target model's scale, this efficiency advantage becomes increasingly pronounced as the size of the target LLM grows.

\subsection{Ablation Study}
\label{subsec:ablation}

To validate the necessity of the \textit{Equivalent Compute} ($C_{\text{pre}}$) mechanism, we conduct ablation studies on Dense-8B and MoE-3B models, comparing our method ($C_{\text{total}} = C_{\text{pre}} + C_{\text{cpt}}$) against two variants:

\begin{itemize}[leftmargin=*]
    \item \textbf{Variant A: Ignoring the Pre-training State ($C_{\text{total}} = C_{\text{cpt}}$).} 
    This variant treats CPT as an independent from-scratch training process, ignoring the model's initial state and prior compute.
    \item \textbf{Variant B: Using Raw Pre-training Tokens.} 
    This variant acknowledges pre-training history but ignores distribution shifts. $C_{\text{pre}}$ is set to the theoretical compute of the original pre-training tokens, typically resulting in a significantly inflated $C_{\text{total}}$.
\end{itemize}

\begin{table}[thbp]
\centering
\small
\renewcommand{\arraystretch}{1.2}
\setlength{\tabcolsep}{4pt} 
\caption{Ablation study on scaling strategies. Scores represent normalized category-averaged performance.}
\label{tab:ablation_variants}
\resizebox{\columnwidth}{!}{%
\begin{tabular}{ll cccc c}
\toprule
\textbf{Model} & \textbf{Strategy ($B, LR$)} 
& \textbf{Math} & \textbf{Know.} & \textbf{Reas.} & \textbf{Code} & \textbf{Avg} \\
\midrule

\multirow{3}{*}{\textbf{Dense-8B}} 
& Variant A ($256, 10^{-4}$)       & $8.04$ & \underline{$9.49$} & \textbf{30.04} & $2.84$ & $50.4$ \\
& Variant B ($1280, 2 \times 10^{-5}$)      & \underline{$8.22$} & \underline{$9.49$} & $29.91$ & \textbf{3.39} & \underline{$51.1$} \\
& Predicted ($768, 5 \times 10^{-5}$)        & \textbf{8.29} & \textbf{9.64} & \textbf{30.04} & \underline{$3.31$} & \textbf{51.3} \\
\midrule

\multirow{3}{*}{\textbf{MoE-3B}} 
& Variant A ($512, 10^{-4}$)       & $11.10$ & $10.88$ & \underline{$28.84$} & $4.34$ & $55.2$ \\
& Variant B ($2560, 2 \times 10^{-5}$)      & \textbf{12.00} & \textbf{11.14} & $28.64$ & \textbf{5.02} & \underline{$56.8$} \\
& Predicted ($1600, 10^{-4}$)       & \underline{$11.70$} & \underline{$11.08$} & \textbf{29.20} & \underline{$4.96$} & \textbf{56.9} \\
\bottomrule
\end{tabular}%
}
\end{table}

As shown in Table~\ref{tab:ablation_variants}, the study reveals two key findings: 
\textbf{(1) Necessity of Accounting for the Initial State.} 
Ignoring $C_{\text{pre}}$ (Variant A) misinterprets CPT as an early-stage training process, leading to an aggressively high learning rate and a smaller batch size. Consequently, the average scores for Dense-8B and MoE-3B drop by $0.9$ and $1.7$ points, respectively. The resulting training instability and initial loss divergence confirm that the pre-trained state must be explicitly accounted for to prevent disrupting the pre-trained weight structure.
\textbf{(2) Superiority of $C_{\text{pre}}$ over Raw Token Counts.} 
Using raw token counts (Variant B) inflates $C_{\text{total}}$, causing the framework to predict an overly conservative $LR$ and a large $B$ by incorrectly assuming the model is near convergence. Our proposed method outperforms Variant B on both Dense-8B ($51.3$ vs. $51.1$) and MoE-3B ($56.9$ vs. $56.8$). This demonstrates that $C_{\text{pre}}$ effectively calibrates the model's knowledge on the target domain, bridging the gap introduced by distribution shifts without requiring costly heuristic tuning.
These results confirm that neither the CPT budget alone nor the original pre-training scale is sufficient for reliable hyperparameter prediction. Instead, $C_{\text{pre}}$ provides a robust metric for determining the optimal optimization trajectory during domain adaptation.

\section{Related Work}

\paragraph{Scaling Laws}
Scaling laws characterize quantitative relationships between model performance and computational resources.
Early work established power-law dependencies of validation loss on model size, dataset size, and compute budget~\cite{kaplan2020scaling,hoffmann2022training}.
Recent studies have extended such laws to hyperparameters, showing that the optimal learning rate follows power laws with respect to model size and data volume, while the optimal batch size is governed by data scale~\cite{li2025steplaw,zhang2025cbs,shuai2024batchsize}.
The critical batch size (CBS) is known to scale with training data volume~\cite{mccandlish2018empirical,zhang2025cbs}.
However, existing hyperparameter scaling laws focus almost exclusively on scratch training, with continued pre-training underexplored.

\paragraph{Continued Pre-training}
Continued pre-training adapts foundation models to target domains while balancing new knowledge acquisition and original capability retention~\cite{gupta2023continual,luo2024empirical}.
Practical techniques such as learning rate rewarming, re-decay, and data replay can help approach the performance of training from scratch~\cite{ibrahim2024simple,parmar2024reuse}.
Warmup-Stable-Decay (WSD) schedules stabilize training and mitigate oscillations in flat loss landscapes~\cite{hu2024minicpm,wen2024wsd,hagele2024scaling}.
Despite these advances, hyperparameter selection (e.g., learning rate, batch size) still relies heavily on heuristic tuning, and a principled, predictive theory for hyperparameter scaling in continued pre-training remains missing.

\paragraph{Hyperparameter Optimization}
Hyperparameter optimization using scaling-based parameterization ($\mu$P) enables zero-shot transfer across model scales via principled parameterization~\cite{yang2023depth,dey2023cerebrasgpt}.
For batch size, gradient-noise-scale-based CBS estimation relies on strong assumptions, while direct measurement through warmup steps offers greater practicality~\cite{mccandlish2018empirical,bergsma2025cbs}.
Recent work also identifies power-law relations among learning rate, batch size, and computation~\cite{shen2024powerscheduler,everett2024scaling}.
Despite progress, two critical gaps remain: existing methods target training from scratch rather than continued pre-training, and $\mu$P enables scale transfer but not dynamic hyperparameter adjustment from a given checkpoint.
To bridge these gaps, we introduce \textit{Equivalent Compute} to quantify the effective computational contribution of prior training, allowing to predict optimal hyperparameters for continued pre-training starting from any given checkpoint.

\section{Conclusion}
In this work, we have presented a novel framework for establishing hyperparameter scaling laws in the context of LLMs continued pre-training. By systematically analyzing the optimization trajectories of small-scale proxy models, we identified predictable relationships between the total compute budget and optimal configurations for learning rate and batch size. To address the challenge of non-zero initialization, we introduced the concept of \textit{Equivalent Compute}, which quantifies a pre-trained model's optimization state relative to a target domain. This formulation enables the direct prediction of optimal hyperparameters for a given checkpoint and compute budgets without requiring any trial runs on the target models. Extensive experiments on models up to 8B parameters demonstrate that our framework significantly improves training stability and final performance while reducing computational search costs by up to 92\%. Overall, our findings offer a principled and scalable methodology for efficient domain adaptation, eliminating the need for exhaustive and costly grid searches.

\newpage
\section*{Limitations}
Despite the effectiveness of our framework, several limitations remain for future investigation:
Verification on Ultra-Large Scales: Due to the immense computational resources required, our empirical validation was primarily conducted on models up to Dense-8B and MoE-3B. While the observed scaling trends are robust at these scales, further research is needed to verify whether specific power-law constants remain invariant or require re-calibration for ultra-large-scale models (e.g., 70B+ parameters), where emergent properties or training instabilities might occur.
Domain-Specific Scaling Sensitivity: Our framework utilizes a unified scaling law fitted on a representative data mixture. However, different data distributions (e.g., formal code vs. mathematical reasoning) may exhibit slightly different scaling exponents ($\gamma$) and irreducible loss floors ($L_0$). While our mixture-based approach demonstrates strong generalizability across the evaluated domains, applying the framework to an extremely specialized or narrow domain might necessitate a localized calibration of the scaling parameters to maintain high prediction accuracy.

\bibliography{custom}

\appendix

\begin{figure*}[t]
    \centering
    \begin{subfigure}[b]{0.48\textwidth}
        \centering
        \includegraphics[width=\linewidth]{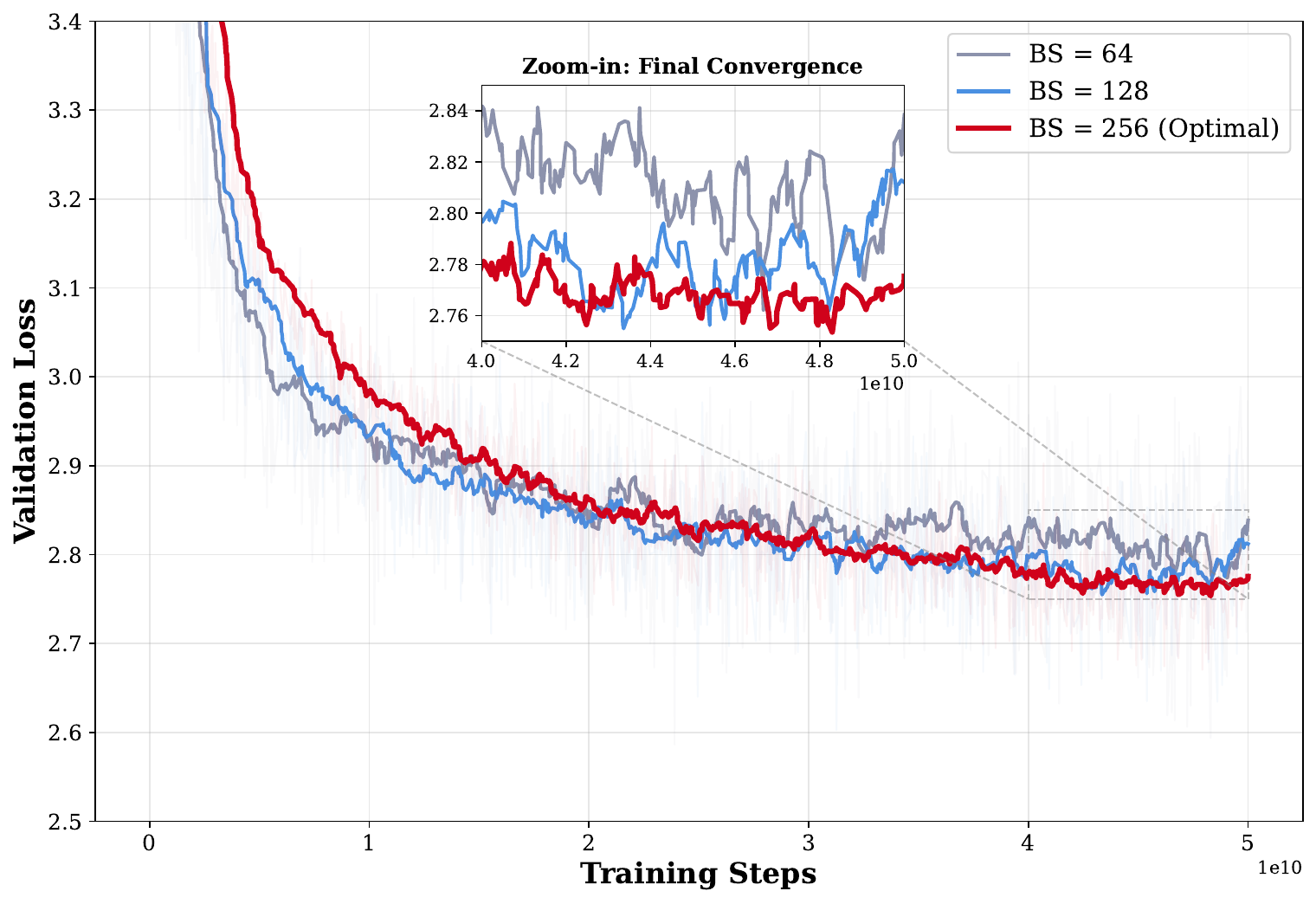}
        \caption{Impact of Batch Size ($B$)}
        \label{fig:sub_batch_size}
    \end{subfigure}
    \hfill
    \begin{subfigure}[b]{0.48\textwidth}
        \centering
        \includegraphics[width=\linewidth]{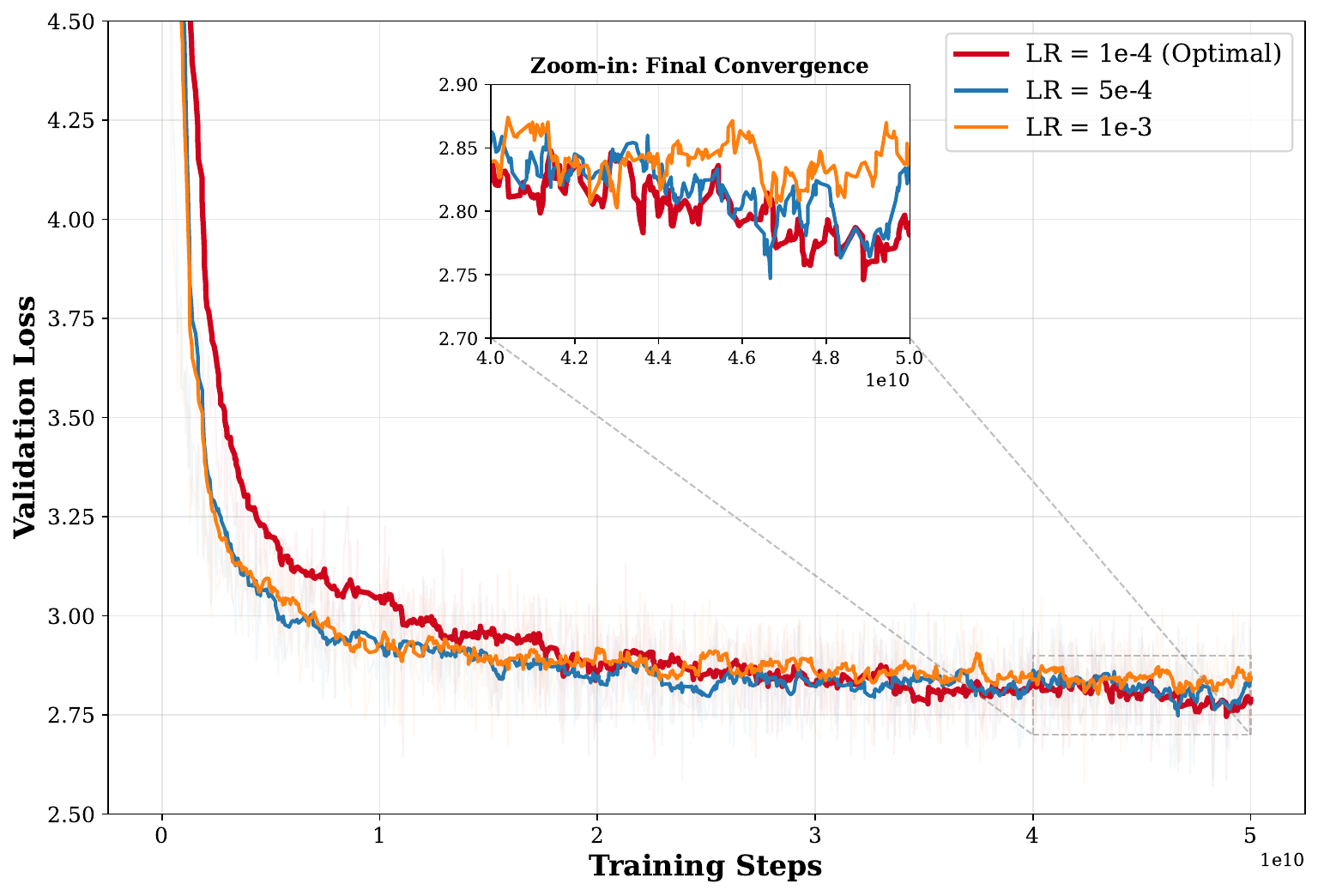}
        \caption{Impact of Learning Rate ($LR$)}
        \label{fig:sub_learning_rate}
    \end{subfigure}
    \caption{Validation loss trajectories during continued pre-training. (a) Larger batch sizes (e.g., $B=256$) enhance gradient stability and yield lower terminal loss, albeit with slower initial reduction. (b) A lower learning rate (e.g., $LR = 1 \times 10^{-4}$) effectively mitigates late-stage loss oscillations, ensuring superior convergence compared to more aggressive schedules.}
    \label{fig:hyperparameter_comparison}
\end{figure*}

\begin{table*}[t]
\centering
\small
\renewcommand{\arraystretch}{1.1}
\setlength{\tabcolsep}{12pt}
\caption{Detailed architectural specifications for Dense-8B and MoE-3B models.}
\label{tab:model_configs}
\begin{tabular}{lclc}
\toprule
\multicolumn{2}{c}{\textbf{Dense-8B Configuration}} & \multicolumn{2}{c}{\textbf{MoE-3B Configuration}} \\
\cmidrule(lr){1-2} \cmidrule(lr){3-4}
\textbf{Parameter} & \textbf{Value} & \textbf{Parameter} & \textbf{Value} \\
\midrule
Vocab Size            & 131,072 & Layers                  & 14 \\
Hidden Size           & 4,096   & Hidden Size             & 3,072 \\
Layers                & 32      & Number of Experts       & 8 of 256 \\
Head Size             & 32      & Dense FFN Hidden Size   & 6,144 \\
Intermediate Size     & 14,336  & Expert FFN Hidden Size  & 1,024 \\
Sequence Length       & 8,192   & Attention Type          & MLA \\
Attention Type        & GQA     & Q-LoRA / KV-LoRA Rank   & 1,536 / 512 \\
Activation            & SwiGLU  & NOPE / RoPE / V-Head    & 128 / 64 / 128 \\
\bottomrule
\end{tabular}
\end{table*}

\section{Model Architecture Details}
\label{appendix:model_details}
This appendix provides the architectural specifications of the models evaluated in our study. Table~\ref{tab:model_configs} summarizes the configurations for both the Dense-8B and MoE-3B models. The MoE-3B model specifically incorporates Multi-Head Latent Attention (MLA) and a sparse Mixture-of-Experts (MoE) layer to optimize compute efficiency.

\section{Empirical Trends of Hyperparameters During Training}
\label{appendix:empirical_trends}

In this section, we examine the empirical evolution of key hyperparameters throughout the training process. Our observations suggest a systematic trend: as training progresses and the model moves toward convergence, achieving optimal performance necessitates a progressively larger batch size and a reduced learning rate.

\paragraph{Trends in Batch Size.} 
Figure~\ref{fig:hyperparameter_comparison}(a) illustrates the training loss trajectories across different batch sizes ($B \in \{64, 128, 256\}$). Initially, the largest batch size ($B = 256$) exhibits a slower rate of loss reduction per token compared to smaller configurations. However, as training progresses, a distinct crossover occurs: the $B = 256$ configuration eventually achieves a lower loss, ultimately converging to the most favorable terminal value. This suggests that while smaller batches provide a stronger stochastic signal early on, larger batches are essential for fine-grained optimization in the later stages.

\paragraph{Trends in Learning Rate.} 
A parallel trend is observed for the learning rate, as shown in Figure~\ref{fig:hyperparameter_comparison}(b). Although the smallest learning rate ($LR = 1 \times 10^{-4}$) initially lags behind more aggressive schedules in terms of reduction speed, it shows superior long-term stability. Eventually, the $LR = 1 \times 10^{-4}$ configuration outperforms its counterparts, attaining a lower final loss. This reinforces the necessity of decaying the learning rate to navigate the flatter regions of the loss landscape as training matures.

\section{Scaling Laws for Optimal Hyperparameters}
\label{appendix:scaling_laws}

\begin{table}[t]
    \centering
    \footnotesize 
    \renewcommand{\arraystretch}{1.1} 
    \setlength{\tabcolsep}{3.5pt}
    \caption{Linear regression parameters for optimal hyperparameters ($y = ax + b$). Here, $x$ denotes the hyperparameter ($\ln(LR)$ or $B$), and $y$ represents  $\ln(\text{FLOPs})$ .}
    \label{tab:scaling_results}
    \begin{tabular}{llccc}
        \toprule
        \textbf{Hypara. ($x$)} & \textbf{Arch.} & \textbf{Slope ($a$)} & \textbf{Intercept ($b$)} \\
        \midrule
        \multirow{2}{*}{$\ln(LR)$} & Dense & $-0.7476$ & $42.1357$ \\
                              & MoE   & $-2.0137$ & $29.9737$ \\
        \midrule
        \multirow{2}{*}{$B$}     & Dense & $0.0093$ & $42.2100$ \\
                              & MoE   & $0.0042$ & $43.1900$ \\
        \bottomrule
    \end{tabular}
\end{table}

To quantitatively analyze the evolution of optimal hyperparameters relative to the computational scale, we perform a linear regression of the optimal hyperparameter values ($x$) against the logarithm of the total compute budget ($y = \ln(\text{FLOPs})$). We model this relationship as $y = ax + b$, where $a$ represents the sensitivity of the compute budget to hyperparameter changes. The resulting scaling parameters and goodness-of-fit ($R^2$) for both Dense and MoE architectures are summarized in Table~\ref{tab:scaling_results}.

\paragraph{Analysis of Learning Rate Scaling.}
The regression analysis reveals divergent scaling trajectories between Dense and MoE architectures, reflecting fundamental differences in their optimization dynamics. For the learning rate, the scaling coefficient $dx/dy$ (calculated as $1/a$) for MoE ($\approx -0.496$) is significantly less negative than that of Dense models ($\approx -1.338$). This indicates that MoE architectures exhibit reduced sensitivity to computational expansion. We hypothesize that this stems from the sparse activation pattern: while the total parameter count scales, the number of active parameters per token grows more conservatively. Consequently, MoE models maintain lower optimization inertia, allowing them to sustain larger, more effective step sizes without the stability risks typically associated with scaling dense networks.

\paragraph{Analysis of Batch Size Scaling.}
Regarding the batch size, the growth rate $dx/dy$ for MoE ($\approx 238.1$) is more than double that of Dense models ($\approx 107.5$). This reveals a disproportionate reliance on batch parallelism in sparse architectures. From a statistical perspective, MoE models require a much larger global batch to ensure sufficient token coverage for each expert and to stabilize the routing mechanism. As the model scales, an aggressive expansion of batch size is mandatory to mitigate the stochastic variance in expert selection and to maintain gradient fidelity across sparse layers. This implies that for future exascale MoE models, the primary scaling bottleneck may shift from raw compute to the efficient utilization of massive batch-level parallelism.

\section{Modeling and Prediction Algorithm}
\label{sec:algorithm}

This appendix provides the formal procedural description of our proposed framework, covering both the cold-start modeling of scaling laws and the zero-shot hyperparameter prediction for a given checkpoint.

\begin{algorithm}[t]
\caption{Hyperparameter Scaling Law Framework for Continued Pre-training}
\label{alg:ohp_framework}
\begin{algorithmic}[1]
\REQUIRE Small-scale proxy models $\mathcal{N}$, target dataset $\mathcal{D}_{\text{cpt}}$, initial pretrained checkpoint $M_{\theta_0}$, planned compute budget $C_{\text{cpt}}$
\ENSURE Predicted optimal hyperparameters $B_{\text{opt}}$ and $LR_{\text{opt}}$

\STATE \textbf{Stage 1: Cold-Start Modeling (Offline)}
\FOR{each model size $N \in \mathcal{N}$}
    \STATE Perform grid search over $(B, LR)$ on $\mathcal{D}_{\text{cpt}}$.
    \STATE Construct the optimal scaling law of $(\text{Loss } L, \text{Compute } C)$ trajectories.
    \STATE Identify optimal configurations $(B_{\text{opt}}, LR_{\text{opt}})$ that minimize $C$ for a given $L$.
\ENDFOR
\STATE Fit mapping functions: $B_{\text{opt}} = f(L)$ and $LR_{\text{opt}} = g(L)$.
\STATE Fit the loss-compute scaling law: $L(C) = L_0 + \alpha C^{-\gamma}$.

\STATE \textbf{Stage 2: Hyperparameter Prediction (Online)}
\STATE $L_{\text{init}} \leftarrow \text{Evaluate}(M_{\theta_0}, \mathcal{D}_{\text{cpt}})$ \COMMENT{Initial validation loss on target data}
\STATE $C_{\text{pre}} \leftarrow \left( \frac{\alpha}{L_{\text{init}} - L_0} \right)^{1/\gamma}$ \COMMENT{Estimate equivalent pre-training compute}
\STATE $C_{\text{total}} \leftarrow C_{\text{pre}} + C_{\text{cpt}}$ \COMMENT{Calculate total effective compute budget}
\STATE $L_{target} \leftarrow L_0 + \alpha \cdot C_{\text{total}}^{-\gamma}$ \COMMENT{Predict target loss at $C_{\text{total}}$}
\RETURN $B_{\text{opt}} = f(L_{target})$, $LR_{\text{opt}} = g(L_{target})$
\end{algorithmic}
\end{algorithm}

\end{document}